\documentclass[10pt,twocolumn,letterpaper]{article}

\usepackage{wacv}
\usepackage{times}
\usepackage{epsfig}
\usepackage{graphicx}
\usepackage{amsmath}
\usepackage{amssymb}

\usepackage[para,online,flushleft]{threeparttable} 
\usepackage{multirow}

\wacvfinalcopy 

\ifwacvfinal
\def\assignedStartPage{1} 
\fi

\ifwacvfinal
\usepackage[breaklinks=true,bookmarks=false]{hyperref}
\else
\usepackage[pagebackref=true,breaklinks=true,colorlinks,bookmarks=false]{hyperref}
\fi

\ifwacvfinal
\else
\fi

\begin{document}

\title{Cross-Modal Virtual Sensing for Combustion Instability Monitoring}

\author{Tryambak Gangopadhyay\\
Iowa State University\\
Ames, IA, USA\\
{\tt\small tryambak@iastate.edu}
\and
Vikram Ramanan\\
IIT Madras\\
Chennai, India\\
{\tt\small vikrambest@yahoo.co.in}
\and
Satyanarayanan R Chakravarthy\\
IIT Madras\\
Chennai, India\\
{\tt\small satyachakra@gmail.com}
\and
Soumik Sarkar\\
Iowa State University\\
Ames, IA, USA\\
{\tt\small soumiks@iastate.edu}
}

\maketitle

\begin{abstract}
   In many cyber-physical systems, imaging can be an important but expensive or `difficult to deploy' sensing modality. One such example is detecting combustion instability using flame images, where deep learning frameworks have demonstrated state-of-the-art performance. The proposed frameworks are also shown to be quite trustworthy such that domain experts can have sufficient confidence to use these models in real systems to prevent unwanted incidents. However, flame imaging is not a common sensing modality in engine combustors today. Therefore, the current roadblock exists on the hardware side regarding the acquisition and processing of high-volume flame images. On the other hand, the acoustic pressure time series is a more feasible modality for data collection in real combustors. To utilize acoustic time series as a sensing modality, we propose a novel cross-modal encoder-decoder architecture that can reconstruct cross-modal visual features from acoustic pressure time series in combustion systems. With the ``distillation" of cross-modal features, the results demonstrate that the detection accuracy can be enhanced using the virtual visual sensing modality. By providing the benefit of cross-modal reconstruction, our framework can prove to be useful in different domains well beyond the power generation and transportation industries.
\end{abstract}

\section{Introduction}

In aerospace and energy industries, ultra-lean premixed combustion is preferred to make gas turbine engines more fuel-efficient with lower cost and low NOx (nitrogen oxides) emissions. With this attempt to make engines efficient and environment-friendly, such operating regimes can make engines more prone to an undesirable phenomenon called combustion instability, which is caused by the establishment of a positive feedback loop between heat release rate fluctuations and fluctuating acoustic pressure \cite{rayleigh1878explanation}.
In confined environments, fluctuating heat release rate leads to the generation of sound waves, which get reflected back to modify the heat release rate. A positive feedback loop can cause a growth of pressure fluctuations leading to large levels of vibration in an engine \cite{culick2006unsteady, dowling1997nonlinear}. This can result in huge revenue loss due to poor performance, reduced life, or catastrophic failure of engine \cite{fisher2009remembering}. Significant trade-offs are required in fuel efficiency and the design of combustion systems to prevent combustion instability (\cite{lieuwen2012unsteady}). To avoid these trade-offs, it is important to develop an accurate and feasible framework for active detection and control of instability.  

Previously, researchers have studied combustion instability using full-scale computational fluid dynamics \cite{palies2011modeling}, physics-based \cite{bellows2007flame} and reduced order \cite{stow2004low} modeling approaches. However, these approaches may require simplifying assumptions and face difficulty in achieving validation. An alternative is to implement data-driven methods utilizing acoustic pressure time-series \cite{gotoda2011dynamic, nair2014multifractality, sen2018dynamic}. These data-driven methods, based only on acoustic pressure time series, can sometimes be inaccurate due to interference from broadband background noise. Recently, researchers have started developing instability detection frameworks using machine learning \cite{kobayashi2019early, sengupta2020bayesian}. With the rapid development in the field of computer vision, the application of deep learning models has started in this domain to detect instability from flame images \cite{sarkar2015early1, akintayo2016prognostics, gangopadhyay2020deep, gangopadhyay2020interpretable, gangopadhyay20213d}.
The effectiveness of model interpretability mechanisms such as `attention’ has been investigated from a domain knowledge perspective \cite{gangopadhyay2020interpretable} and deep learning results have been verified from a physics-based understanding \cite{gangopadhyay20213d}. Therefore, the image-based deep learning frameworks have proved to be accurate, trustworthy and can build the confidence of domain experts to implement these models in real systems to detect combustion instability. However, the current roadblock exists on the hardware side. 

Acquisition and processing of high-volume flame image data may not be feasible to perform fast enough using existing commercial hardware. Also, flame imaging is not a common sensing modality in engines today. Therefore, the image-based deep learning frameworks can only become feasible with rapid improvement in the hardware sector. Acoustic pressure time series is a more feasible modality for data collection in real combustors. To circumvent the hardware roadblock and simultaneously ensure high detection accuracy, the optimal solution can be to utilize acoustic time series as a sensing modality and implement image-based deep learning models. In this work, we attempt to think in that direction by proposing a novel virtual sensing model (VSenseNet) to reconstruct cross-modal visual features from acoustic pressure time series in combustion systems. While researchers have proposed cross-modal reconstruction models for text-to-image \cite{reed2016generative, xu2018attngan, li2019object}, and speech-to-face \cite{oh2019speech2face, duarte2019wav2pix}, there has been no work on the reconstruction of visual features from time series in any application domain.

\textbf{Contributions.} We summarize the contributions of this work as follows:

\begin{enumerate}

    \item To the best of our knowledge, this is the first work on cross-modal reconstruction of visual features from time series in any domain. The proposed cross-modal encoder-decoder model VSenseNet is novel in the context of combustion systems to reconstruct flame images from acoustic pressure time series. 
    
    \item In VSenseNet, visual reconstruction from time series is achieved by training the encoder-decoder with ``distillation" of cross-modal features from models pre-trained on images. During testing, the classification performance of synthetic images is compared against that of actual images.
    
    \item With acoustic time series as the sensing modality, we demonstrate that instability detection accuracy can be enhanced using our proposed virtual sensing modeling approach. VSenseNet can prove to be a great resource in different sectors where imaging is an important but `difficult to deploy' sensing modality.

\end{enumerate}

\section{Related Work}

Researchers have proposed models involving cross-modal reconstruction for text, speech, and image datasets. Conditional Generative Adversarial Nets (GANs) \cite{mirza2014conditional} can direct the data generation process by conditioning the model on additional information. 
A training strategy involving GAN \cite{goodfellow2014generative} architecture can enable text-to-image synthesis of bird and flower images from human-written descriptions \cite{reed2016generative}.
Conditional GANs have been used to achieve the cross-modal audio-visual generation of musical performances \cite{chen2017deep}.
AttnGAN \cite{xu2018attngan} is an attention-driven model for text-to-image generation where the layered attentional structure can pay attention to the relevant words in the natural language description for generating different parts of the image. 
Obj-GAN \cite{li2019object} is an object-driven attentive generative network for synthesizing complex images from text descriptions utilizing an object-driven attentive generative network and an object-driven discriminator.
Speech2Face \cite{oh2019speech2face} model has been proposed to study the task of reconstructing a facial image of a person from a short audio recording of that person speaking.
Face images of a speaker can be generated with a self-supervised approach by exploiting the audio and visual signals naturally aligned in videos \cite{duarte2019wav2pix}.
Cross-modal matching can be used to generate faces from voices that match several biometric characteristics of the speaker \cite{wen2019face}.
A model has been proposed to use both audio and a low-resolution image to perform extreme face super-resolution \cite{meishvili2020learning}.


\section{Virtual Sensing Model (VSenseNet)}

To address the hardware roadblock of flame image acquisition, to use acoustic time series as sensing modality, and to simultaneously utilize an image-based detection framework for better accuracy, we propose a novel cross-modal encoder-decoder virtual sensing model VSenseNet. In this section, we demonstrate two versions of VSenseNet - VSenseNet I and VSenseNet II. The training framework of VSenseNet II additionally comprises of image classifier, while that of VSenseNet I only includes the time series encoder and the image decoder.
We demonstrate the ablation studies of both versions in the supplementary materials.

\subsection{Convolutional Autoencoder Pre-Training}

\begin{figure*}
    \centering
    \includegraphics[width=12cm]{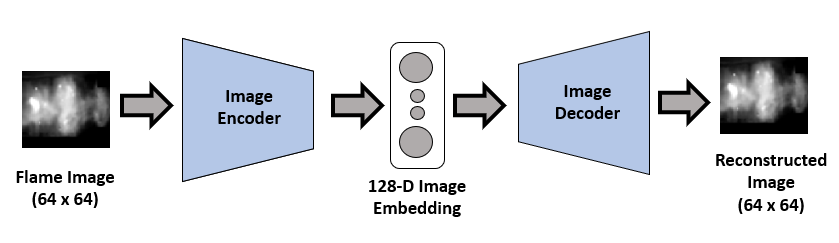}
    \caption{The encoder and decoder of the convolutional autoencoder model used for pre-training.}
    \label{autoencoder}
\end{figure*}

Autoencoders can learn meaningful representations using a compression function (encoder) and a decompression function (decoder). The encoder compresses the input into a low dimensional embedding, and the decoder reconstructs the high dimensional information from that embedding.
Without the requirement of explicit annotations, the weights of an autoencoder model can be learned with the objective of minimizing the reconstruction loss.
The first step is to utilize the training dataset of flame images to pre-train a convolutional autoencoder which comprises an image encoder and an image decoder as demonstrated in Fig.~\ref{autoencoder}.
The encoder takes in a flame image (resolution 64 x 64) as input. The encoder model comprises a series of 2D convolutional and 2D max-pooling layers. After that, a fully connected layer is used to compute a 128-dimensional embedding. From the 128-dimensional embedding, the decoder attempts to reconstruct the original flame image as closely as possible using a series of 2D up-sampling and 2D convolutional transpose layers. The details of image encoder and decoder models are provided in the supplementary section.

\subsection{Time Series Encoder}

\begin{figure}
    \begin{center}
    \includegraphics[width=8cm]{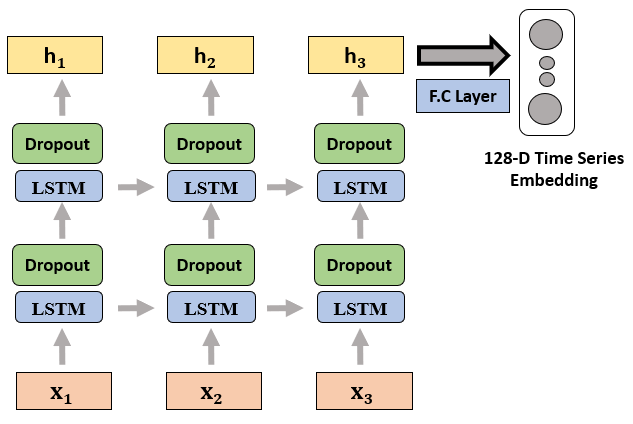}
    \end{center}
\caption{The details of time series encoder for training the VSenseNet.}
\label{ts_encoder}
\end{figure}

The training frameworks of both versions of VSenseNet consist of the time series encoder to compute an embedding from the time series.
Long Short Term Memory (LSTM) networks can effectively capture long-term temporal dependencies \cite{hochreiter1997long}, and LSTM networks are effective in different applications involving time series data \cite{ gangopadhyay2021spatiotemporal}.
We develop the time series encoder model consisting of two LSTM layers with dropout added after each layer to prevent over-fitting. The time series encoder is shown in Fig.~\ref{ts_encoder}.
The first LSTM layer takes in the multivariate time series as input. The hidden state outputs of the first LSTM layer act as inputs to the second LSTM layer. The last hidden state of the second LSTM layer is considered the compressed information for the time series. A fully connected layer is used to get a 128-dimensional time series embedding.


\subsection{VSenseNet I}

\begin{figure*}
    \centering
    \includegraphics[width=16cm]{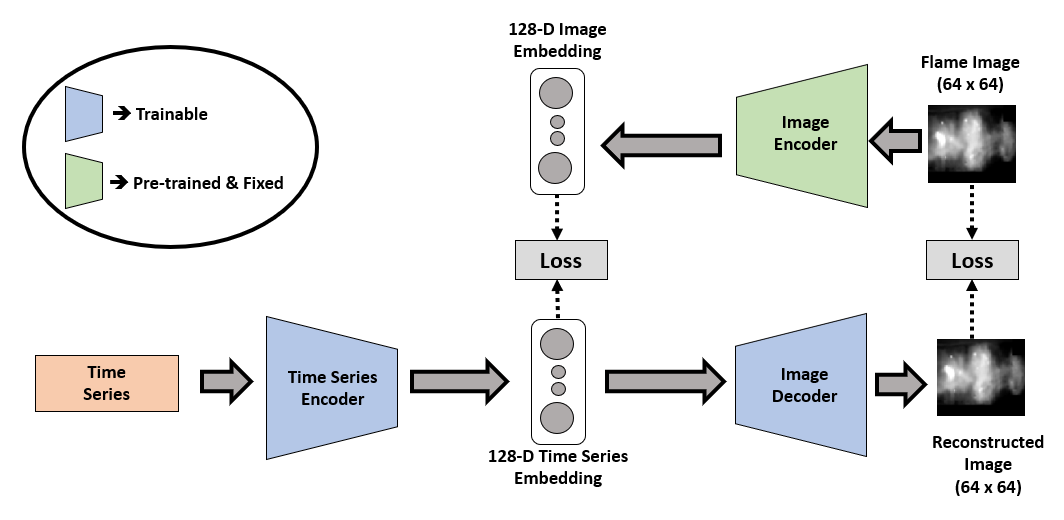}
    \caption{Training framework of VSenseNet I. It utilizes the pre-trained image encoder to compute the image embedding. The time series encoder and image decoder are trained with two loss functions - embedding loss and reconstruction loss.}
    \label{vsensenet_1_training}
\end{figure*}

The VSenseNet I modeling approach is illustrated in Fig.~\ref{vsensenet_1_training}. The trainable parts of the VSenseNet I framework are the time series encoder and image decoder, and the pre-trained (and fixed) part is the image encoder. The weights of VSenseNet I are learned with two learning objectives - minimizing the embedding loss and minimizing the reconstruction loss.

The pre-trained image encoder is utilized to compute the 128-dimensional image embedding. 
The trainable time series encoder (Fig.~\ref{ts_encoder}) generates the 128-dimensional time series embedding. The embedding loss is the mean squared error (MSE) computed between the time series embedding and image embedding. In the training process, the time series encoder learns to compute an embedding that can match the image embedding as closely as possible.

The image decoder in Fig.~\ref{vsensenet_1_training} is trained from scratch alongside the time series encoder in the training loop. The model architecture of the decoder is the same as that in the autoencoder model (Fig.~\ref{autoencoder}). The input to the image decoder is the time series embedding, from which it attempts to reconstruct the image corresponding to the input time series. The learning objective of the image decoder is to minimize the reconstruction loss between the actual flame image and the reconstructed image.

\subsection{VSenseNet II}

\begin{figure*}
    \centering
    \includegraphics[width=17cm]{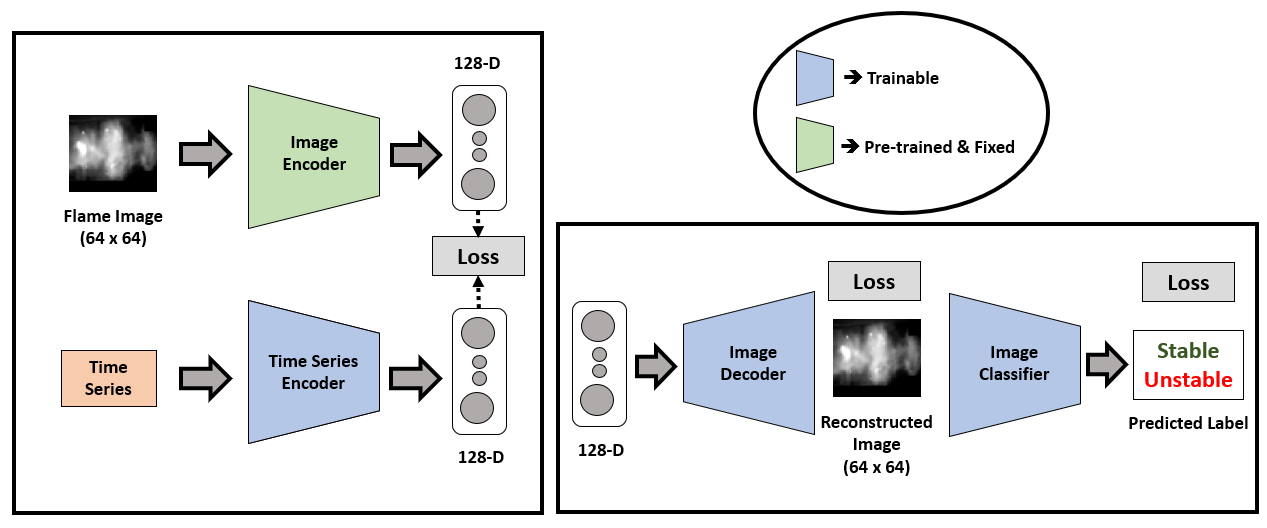}
    \caption{Training framework of VSenseNet II. It comprises two steps. The first step is to train the time series encoder. The next step is to train the image decoder and image classifier utilizing reconstruction loss and classification loss.}
    \label{vsensenet_2_training}
\end{figure*}

Compared to VSenseNet I, VSenseNet II has an image classifier model in the training framework apart from the time series encoder and the image decoder. We demonstrate the training framework of VSenseNet II in Fig.~\ref{vsensenet_2_training}. The trainable parts are the time series encoder, image decoder, and image classifier. Similar to VSenseNet I, in VsenseNet II also, the pre-trained image encoder is utilized. The training framework consists of two steps.

In the first step, the time series encoder is trained to regress to the image embedding computed from the image encoder. With the time series as input, the time series encoder computes a 128-dimensional time series embedding. The learning objective is to minimize the embedding loss (MSE) between the time series and image embeddings.

In the next step, the image decoder and image classifier models are trained. The model architecture of the decoder is the same as that used in Fig.~\ref{autoencoder}. The image classifier model comprises 2D convolutional, 2D max-pooling, and fully connected layers. It is a binary classification model to predict a flame image as stable or unstable. The model architecture of the image classifier is provided in supplementary materials.
The generated time series embeddings are utilized for training this part of the framework. From an embedding, the image decoder learns to reconstruct the corresponding flame image as closely as possible. With the reconstructed flame image as input, the image classifier model predicts it as stable or unstable.

\subsection{Test Framework for VSenseNet}

\begin{figure*}
    \centering
    \includegraphics[width=15cm]{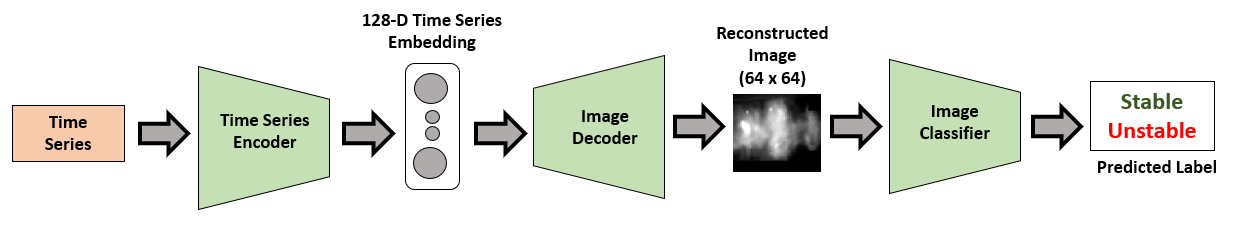}
    \caption{Test framework of VSenseNet comprising time series encoder, image decoder and image classifier.}
    \label{test_framework}
\end{figure*}

The overall test framework for VSenseNet is shown in Fig.~\ref{test_framework}. It consists of three steps - time series encoder, image decoder, and image classifier. 

For VSenseNet I, the time series encoder and the image decoder are trained as shown in Fig.~\ref{vsensenet_1_training}. The image classifier is trained separately using actual flame images of the training dataset. For VSenseNet II, the time series encoder, the image decoder, and the image classifier are trained as shown in Fig.~\ref{vsensenet_2_training}.

After taking in the multivariate time series as input, the trained time series encoder model computes the time series embedding. From this embedding, the trained image decoder computes the reconstructed image. This image is fed into the trained image classifier model to classify it as stable or unstable. Therefore, the virtual sensing framework utilizes acoustic time series as the sensing modality from which it reconstructs the cross-modal visual features. The reconstructed image is then used to identify the presence of combustion instability. With this approach of using a feasible sensing modality, the hardware roadblock of flame image acquisition can be avoided, and simultaneously, better detection accuracy can be achieved.

\section{Experiments}

\subsection{Dataset}

For dataset collection, we induce combustion instability in a laboratory-scale swirl combustor (details provided in the supplementary section). 
The fuel is injected co-axially with air at selected upstream distances. 
For dataset collection, the chosen upstream distances are 90 mm and 120 mm. For the upstream distance of 90 mm, partial premixing of the fuel with air occurs, while the distance of 120 mm facilitates full premixing of the fuel and air.



The ground truth labels (stable, unstable) for the hi-speed flame image sequences are provided by the domain experts.
The conditions are defined as stable or unstable based on the dominant frequency, and root mean square (RMS) value of pressure fluctuations (between initial and final instants).
The stable conditions demonstrate broadband frequency (estimated from fast Fourier transform) and RMS pressure values less than 100 Pa.
For unstable conditions, the frequency of oscillations corresponds to sharp values in the range of 130-150 Hz, and the RMS pressure exceeds 500 Pa.
Hence, the images are labeled using the corresponding pressure modality and not using image features.

We identify the conditions by upstream distance (premixing length), airflow rate (AFR), and fuel flow rate (FFR).
Both AFR and FFR are expressed in lpm (liters per minute).
The hi-speed images are captured at 3000 Hz (with a resolution of 1024 x 1024) for 3 seconds at each condition. 
Simultaneously, the pressure data is recorded at 4 locations of the experimental setup with a frequency of 9000 Hz.
Therefore, for each condition, we have 9000 frames and 27000 time steps of multivariate pressure data. 
From a total of six conditions, we use four conditions for training our proposed models and keep two conditions for testing the performance of the models. 

The stable and unstable conditions in the training set are:

\begin{enumerate}

    \item $\mathbf{Stable_{120/60/600}}$: Condition has Premixing Length = 120 mm, FFR = 60 and AFR = 600.
    
    \item $\mathbf{Stable_{90/45/450}}$: Condition has Premixing Length = 90 mm, FFR = 45 and AFR = 450.
    
    \item $\mathbf{Unstable_{120/45/900}}$: Condition has Premixing Length = 120 mm, FFR = 45 and AFR = 900.
    
    \item $\mathbf{Unstable_{90/28/600}}$: Condition has Premixing Length = 90 mm, FFR = 28 and AFR = 600.
    
\end{enumerate}

The stable and unstable conditions in the test set are:

\begin{enumerate}

    \item $\mathbf{Stable_{120/45/450}}$: Condition has Premixing Length = 120 mm, FFR = 45 and AFR = 450.
    
    \item $\mathbf{Unstable_{90/45/900}}$: Condition has Premixing Length = 90 mm, FFR = 45 and AFR = 900.

\end{enumerate}

\subsection{Baseline Models}

\begin{figure*}
    \centering
    \includegraphics[width=16cm]{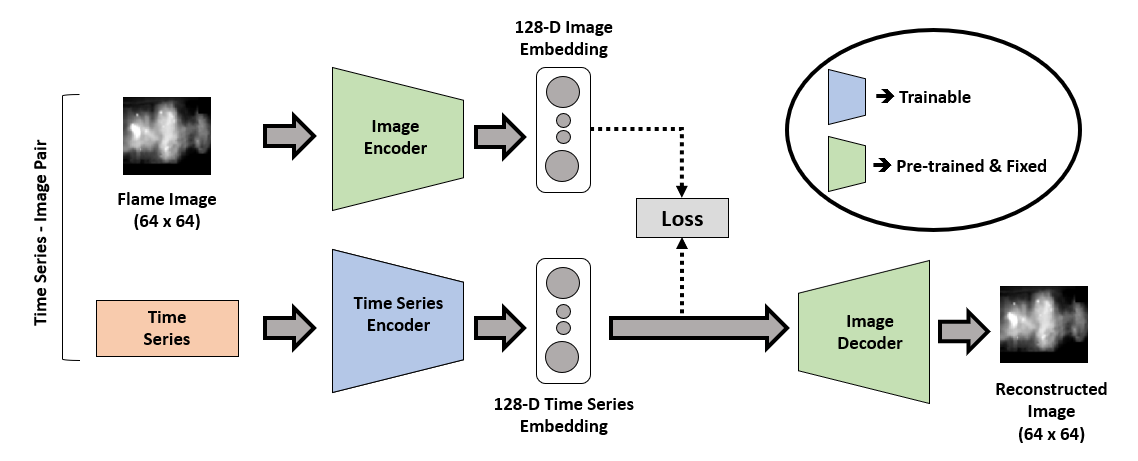}
    \caption{Framework to train the time series encoder of the baseline Cross-Modal Model which is inspired from Speech2Face \cite{oh2019speech2face}.}
    \label{cross_modal_baseline}
\end{figure*}

For comparison of results, we use three baseline models.

\begin{enumerate}

    \item \textbf{Image Classifier:} 
    The image classifier model is the same as that used in the test framework of VSenseNet (Fig.~\ref{test_framework}). This image-based baseline model takes a flame image as input and classifies it as stable or unstable. It is trained on actual training set images and tested on actual test set images.
    
    \item \textbf{Time Series Classifier:}
    This is an entirely time series based model with no cross-modal reconstruction of visual features. The time series classifier model is developed by augmenting the time series encoder model (Fig.~\ref{ts_encoder}) - two fully connected layers are added after the computation of the 128-dimensional time series embedding. The entire model architecture of the time series classifier has been provided in the supplementary section. With the multivariate time series as input, the model predicts it as stable or unstable. The time series classifier model is trained using time series only, and it is tested using time series of the test set.
    
    \item \textbf{Cross-Modal Model:}
    We develop this cross-modal baseline model inspired from the Speech2Face \cite{oh2019speech2face} model. Speech2Face was proposed to reconstruct a human face from an audio recording of a person speaking. Speech2Face model consists of a voice encoder that takes spectrogram as input - the spectrogram is computed from the audio recording. The voice encoder is trained utilizing a pre-trained face encoder network. The voice encoder computes an embedding that is fed to a pre-trained face decoder model to reconstruct the human face. We implement a framework similar to that of Speech2Face - time series encoder is used instead of voice encoder to encode the time series, pre-trained image encoder, and image decoder models (Fig.~\ref{autoencoder}) are used as replacements of pre-trained face encoder and face decoder models, respectively. The framework to train the time series encoder is shown in Fig.~\ref{cross_modal_baseline}. The test framework of the Cross-Modal model is the same as that of VSenseNet, as demonstrated in Fig.~\ref{test_framework}.

\end{enumerate}

\subsection{Results}

In this section, we discuss the classification and reconstruction performance of the proposed approach. 

{
\setlength{\tabcolsep}{6pt}
\begin{table*}[h]
\begin{center}
    \begin{tabular}{|c||c|c||c|c|c|}
    \hline
    \multirow{2}{*}{Model} &
    \multicolumn{2}{c||}{Reconstruction Performance} & \multicolumn{3}{c|}{Classification Performance} \\ 
    \cline{2-3} \cline{4-5} \cline{5-6}
    & SSIM & Mean Squared Error & Accuracy & F1 Score & FNR\\
    \hline
    \hline
    Image Classifier & NA & NA & \textbf{0.9938 $\pm$ 0.0064} & \textbf{0.9938 $\pm$ 0.0065} & \textbf{0.0122 $\pm$ 0.0129}\\
    \hline
    Time Series Classifier & NA & NA & \textbf{0.9850 $\pm$ 0.0021} & \textbf{0.9847 $\pm$ 0.0022} & \textbf{0.0300 $\pm$ 0.0044}\\
    \hline
    \hline
    Cross-Modal & 0.6655 & 0.0072 & 0.9888 $\pm$ 0.0005 & 0.9887 $\pm$ 0.0005 & 0.0222 $\pm$ 0.0010\\
    \hline
    VSenseNet I & \textbf{0.6980} & \textbf{0.0062} & 0.9895 $\pm$ 0.0003 & 0.9894 $\pm$ 0.0002 & 0.0209 $\pm$ 0.0006\\
    \hline
    VSenseNet II & 0.6876 & 0.0070 & \textbf{0.9901 $\pm$ 0.0003} & \textbf{0.9900 $\pm$ 0.0003} & \textbf{0.0197 $\pm$ 0.0007}\\
    \hline
    \end{tabular}
\end{center}
\caption{Empirical Results for the test set. For classification performance, average and standard deviation of the evaluation metrics (Accuracy, F1 Score, FNR) are reported after training each model five times. For reconstruction performance, average values of SSIM and MSE are reported.
}
\label{empirical_results}
\end{table*}
}

\begin{figure*}
    \centering
    \includegraphics[width=15cm]{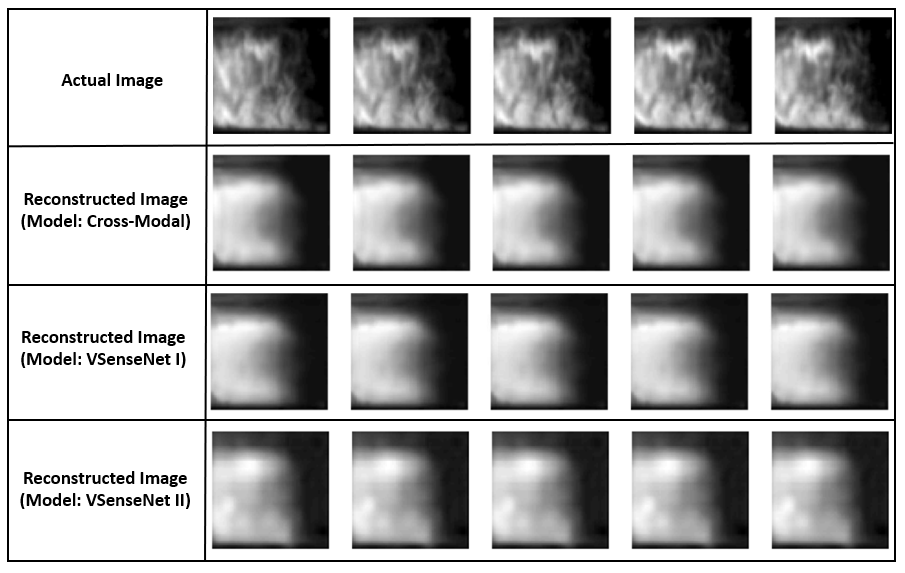}
    \caption{The reconstructions obtained from the proposed VSenseNet models are compared against the reconstructions obtained from the Cross-Modal baseline model, and the actual flame images for the test set $\mathbf{Stable_{120/45/450}}$ condition.}
    \label{test_stable}
\end{figure*}

\begin{figure*}
    \centering
    \includegraphics[width=15cm]{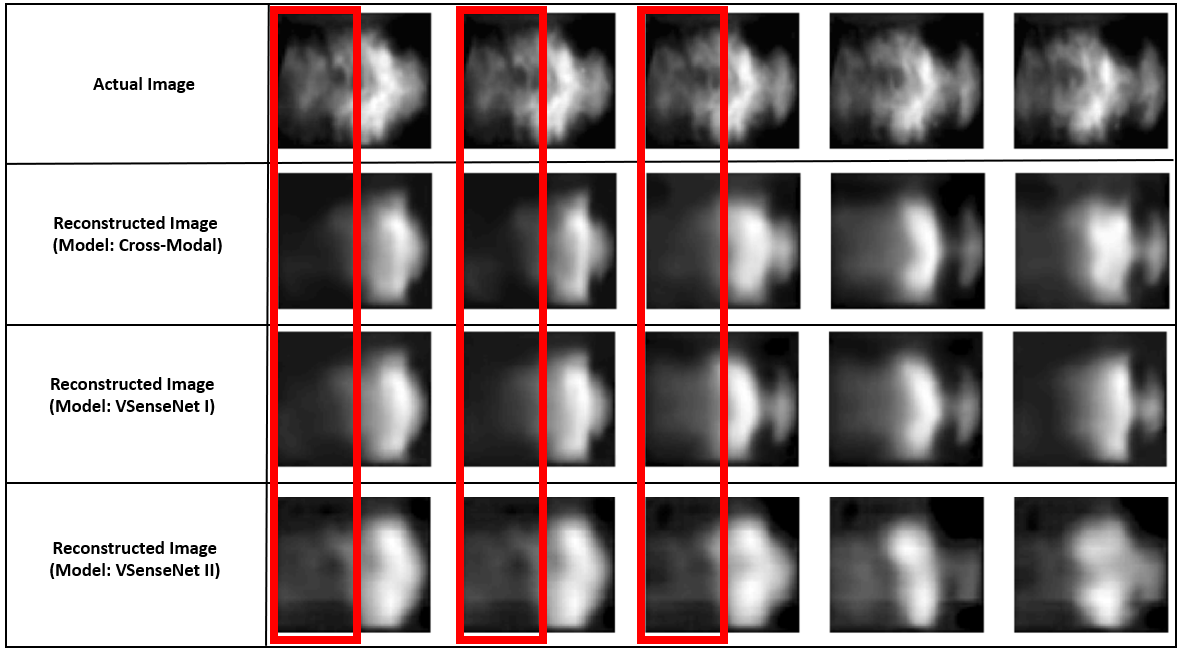}
    \caption{The reconstructions obtained from the proposed VSenseNet models are compared against the reconstructions obtained from the Cross-Modal baseline model, and the actual flame images for the test set $\mathbf{Unstable_{90/45/900}}$ condition.}
    \label{test_unstable}
\end{figure*}

We use three evaluation metrics for classification performance: Accuracy, F1 Score, and False Negative Rate. F1 Score, also known as F-score or F-measure, is the harmonic mean of precision and recall. False negative rate (FNR) refers to falsely predicting negative (stable) when it is actually positive (unstable). FNR is the ratio between the predicted number of false negative samples and the actual number of positive samples. It summarizes how often stable is predicted when the actual is unstable. For detection of combustion instability, it is highly significant to have a low model FNR.
We use Adam optimizer with a learning rate of 0.001 and a batch size of 32. The models are trained using NVIDIA Titan RTX GPU. 

For reconstruction performance, we use two evaluation metrics - Mean Squared Error (MSE) and Structural Similarity Index Measure (SSIM). MSE computed between the actual and reconstructed images may not always be highly indicative of the structural similarity. SSIM \cite{wang2009mean, wang2004image} addresses this issue by considering texture and also including luminance masking and contrast masking terms. Therefore, SSIM can be efficient in estimating the perceived similarity between the actual and reconstructed images. 

Table~\ref{empirical_results} presents the empirical results for the test set conditions. The Image Classifier Model, which is tested using actual flame images, shows an average accuracy of 99.38\%. The Time Series Classifier Model, which doesn't involve any cross-modal reconstruction, shows an average accuracy of 98.50\%. Therefore, the image-based model is more accurate than the time series based framework. 
From Table~\ref{empirical_results}, we observe that our proposed models (VSenseNet I and VSenseNet II) demonstrate better accuracy than the time series model. Both of our proposed models also outperform the Cross-Modal baseline model in terms of accuracy, F1 Score, and FNR. Using time series as the input modality with virtual sensing approach, we can enhance the average classification accuracy from 98.50\% to 99.01\% and approach closer towards the 99.38\% accuracy achieved by the imaging modality-based model.  Therefore, by adopting the proposed training approach of distillation of cross-modal features and reconstruction of synthetic images, we enhance the instability detection performance with acoustic time series as the sensing modality.  

In terms of reconstruction performance, both VSenseNet I and VSenseNet II outperform the Cross-Modal baseline model in terms of SSIM and MSE as demonstrated in Table~\ref{empirical_results}. Sample reconstruction results are shown in Fig.~\ref{test_stable} and Fig.~\ref{test_unstable}.
The reconstructed images are generally a bit more smooth than the actual images.
From Fig.~\ref{test_stable}, we observe that all the models can reliably reconstruct the flame images for the test set $\mathbf{Stable_{120/45/450}}$ condition. For the other test set condition $\mathbf{Unstable_{90/45/900}}$, the proposed VSenseNet models reconstruct the flame structures better than the baseline Cross-Modal model as highlighted by red boxes in Fig.~\ref{test_unstable}.

While stable-unstable flame classification can help in designing active combustion control mechanisms to mitigate the instability, it does not provide any scientific insight to the domain experts in terms of the coherent structures responsible for triggering the instability, and hence, the overall approach may lack interpretability. Therefore, we stress upon the need for flame image reconstruction that can provide valuable insights during offline analysis as well as build sufficient trust of the domain experts via necessary interpretability.

\section{Conclusion}

Deep learning frameworks have demonstrated state-of-the-art performance in detecting combustion instability from flame images. Such frameworks have also proved to be trustworthy to build the confidence of domain experts. But the current roadblock exists in the acquisition of high-volume flame images within the confines of an engine having high temperatures. 
From the hardware side, capturing acoustic pressure time series in real combustors is a more feasible modality.
To utilize acoustic time series as a sensing modality and, at the same time, to simultaneously ensure high detection accuracy, we propose a novel cross-modal encoder-decoder virtual sensing model that can reconstruct cross-modal visual features from acoustic pressure time series in combustion systems. 

Our proposed VSenseNet approach demonstrates effectiveness in reconstructing the flame images.
By choosing different conditions in the test set, we demonstrate the robustness of our model. 
We demonstrate that by cross-modal reconstruction of synthetic images, the classification performance is better than that from time series alone. Therefore we are enhancing the accuracy of combustion instability detection using time series data with our virtual sensing modeling approach.
Domain experts can also gain valuable insights during an offline analysis of the reconstructed flame images. 

VSenseNet provides the unique benefit of generating synthetic visual features corresponding to time series information. 
We believe that our proposed approach has the potential to impact different sectors dealing with cyber-physical systems where imaging is an important but `difficult to deploy' sensing modality.
Our VSenseNet approach can fill up the gap by providing the benefit of cross-modal reconstruction, and we envision that this can bring a transformative advancement in different application domains.
In the future, we plan to extend VSenseNet to capture transitions in a combustion system. We would also like to apply our modeling approach to other cyber-physical systems.





{\small
\bibliographystyle{ieee_fullname}
\bibliography{egbib}
}

\vfill\pagebreak

\section*{Supplementary Materials}

\section*{S.1 Ablation Study}

The training framework of VSenseNet II additionally comprises of image classifier, while that of VSenseNet I only includes the time series encoder and the image decoder. For the ablation study of VSenseNet I and VSenseNet II, we conduct multiple experiments.

\subsection*{S.1.1 VSenseNet I}

\begin{figure*}
    \centering
    \includegraphics[width=16cm]{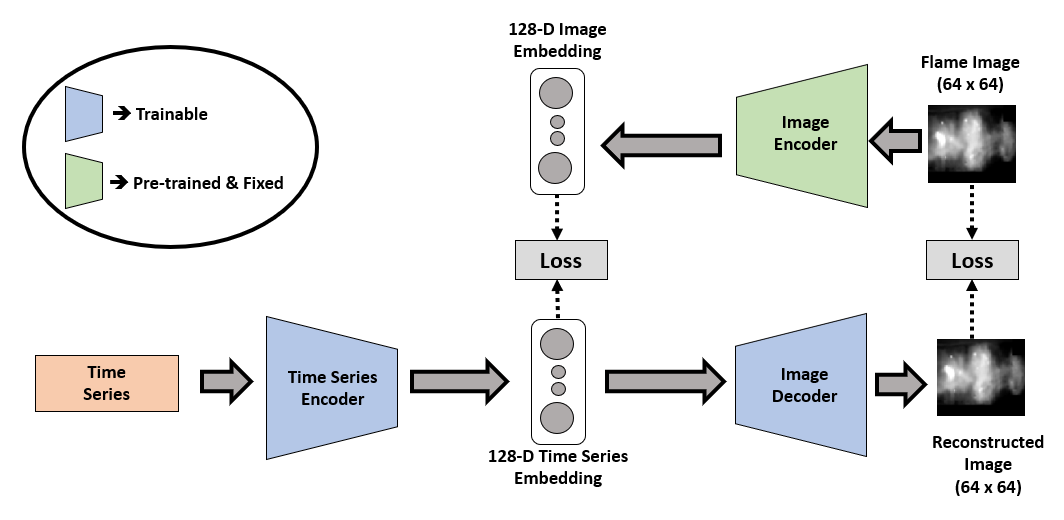}
    \caption{Training framework of VSenseNet I. It utilizes the pre-trained image encoder to compute the image embedding. The time series encoder and image decoder are trained with two loss functions - embedding loss and reconstruction loss.}
    \label{vsensenet_1_training}
\end{figure*}

\begin{figure*}
    \centering
    \includegraphics[width=16cm]{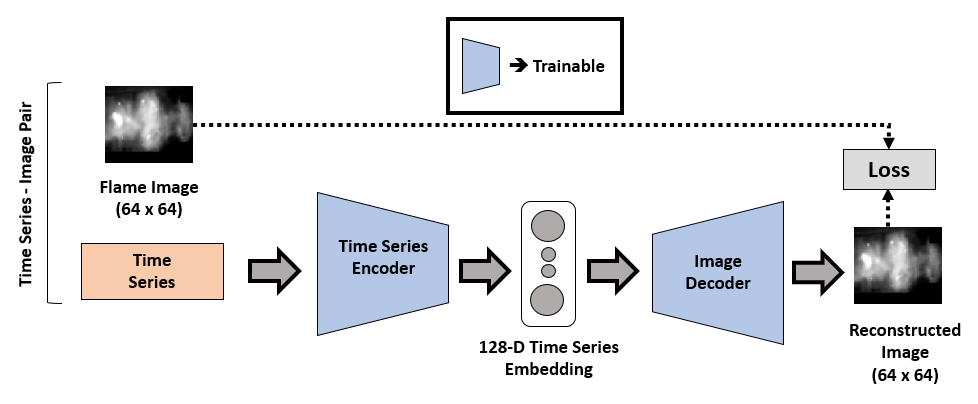}
    \caption{Training framework of VSenseNet I(A). The time series encoder and image decoder are trained with reconstruction loss.}
    \label{vsensenet_1A_training}
\end{figure*}

\begin{figure*}
    \centering
    \includegraphics[width=16cm]{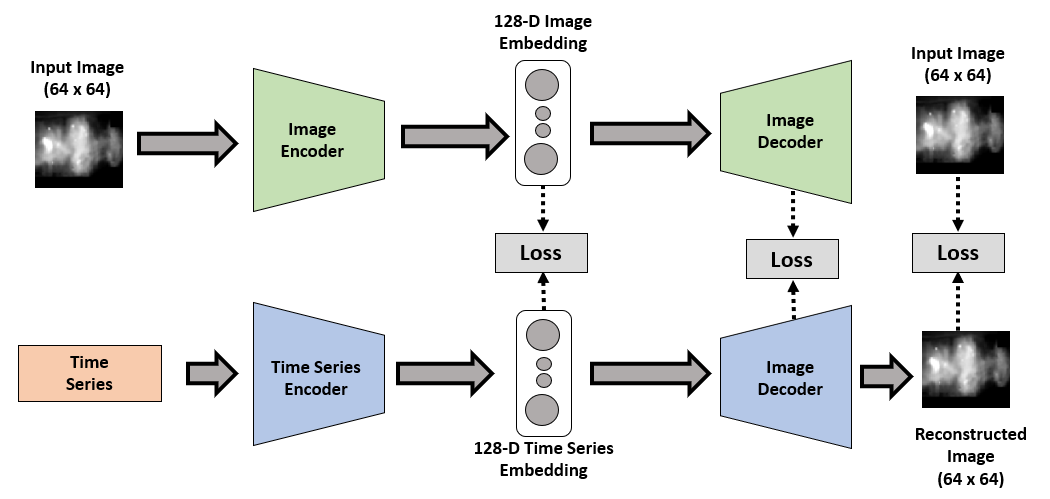}
    \caption{Training framework of VSenseNet I(B). The time series encoder and image decoder are trained with three loss functions.}
    \label{vsensenet_1B_training}
\end{figure*}

The VSenseNet I modeling approach is illustrated in Fig.~\ref{vsensenet_1_training}. The trainable parts of the VSenseNet I framework are the time series encoder and image decoder, and the pre-trained (and fixed) part is the image encoder. The weights of VSenseNet I are learned with two learning objectives - minimizing the embedding loss and minimizing the reconstruction loss.
The pre-trained image encoder is utilized to compute the 128-dimensional image embedding. 
The trainable time series encoder generates the 128-dimensional time series embedding. The embedding loss is the mean squared error (MSE) computed between the time series embedding and image embedding. In the training process, the time series encoder learns to compute an embedding that can match the image embedding as closely as possible.
The input to the image decoder is the time series embedding, from which it attempts to reconstruct the image corresponding to the input time series. The learning objective of the image decoder is to minimize the reconstruction loss between the actual flame image and the reconstructed image.

As part of ablation study for VSenseNet I, we develop the models VSenseNet I(A) and VSenseNet I(B), demonstrated in Fig.~\ref{vsensenet_1A_training} and Fig.~\ref{vsensenet_1B_training} respectively. In VSenseNet I(A), we remove the embedding loss - the time series encoder and the image decoder are trained with only reconstruction loss. In VSenseNet I(B), we add another loss function in addition to the embedding loss and reconstruction loss. The additional loss is added after the second convolutional transpose layer of the image decoder to facilitate knowledge distillation from the image decoder pre-trained on images. From Table~\ref{ablation_study_vsensenet_1}, we observe that while VSenseNet I(A) is better than VSenseNet I(B) in terms of reconstruction performance, VSenseNet I(B) outperforms VSenseNet I(A) in terms of classification performance.
Overall, VSenseNet I performs better in terms of both reconstruction and classification performance.

{
\setlength{\tabcolsep}{6pt}
\begin{table*}[h]
\begin{center}
    \begin{tabular}{|c||c|c||c|c|c|}
    \hline
    \multirow{2}{*}{Model} &
    \multicolumn{2}{c||}{Reconstruction Performance} & \multicolumn{3}{c|}{Classification Performance} \\ 
    \cline{2-3} \cline{4-5} \cline{5-6}
    & SSIM & Mean Squared Error & Accuracy & F1 Score & FNR\\
    \hline
    \hline
    Image Classifier & NA & NA & \textbf{0.9938 $\pm$ 0.0064} & \textbf{0.9938 $\pm$ 0.0065} & \textbf{0.0122 $\pm$ 0.0129}\\
    \hline
    Time Series Classifier & NA & NA & \textbf{0.9850 $\pm$ 0.0021} & \textbf{0.9847 $\pm$ 0.0022} & \textbf{0.0300 $\pm$ 0.0044}\\
    \hline
    \hline
    Cross-Modal & 0.6655 & 0.0072 & 0.9888 $\pm$ 0.0005 & 0.9887 $\pm$ 0.0005 & 0.0222 $\pm$ 0.0010\\
    \hline
    VSenseNet I & \textbf{0.6980} & 0.0062 & \textbf{0.9895 $\pm$ 0.0003} & \textbf{0.9894 $\pm$ 0.0002} & \textbf{0.0209 $\pm$ 0.0006}\\
    \hline
    VSenseNet I(A) & 0.6965 & \textbf{0.0061} & 0.9859 $\pm$ 0.0011 & 0.9857 $\pm$ 0.0012 & 0.0280 $\pm$ 0.0023\\
    \hline
    VSenseNet I(B) & 0.6894 & 0.0063 & 0.9890 $\pm$ 0.0002 & 0.9889 $\pm$ 0.0002 & 0.0218 $\pm$ 0.0005\\
    \hline
    \end{tabular}
\end{center}
\caption{Ablation study with VSenseNet I for the test set. For classification performance, average and standard deviation of the evaluation metrics (Accuracy, F1 Score, FNR) are reported after training each model five times. For reconstruction performance, average values of SSIM and MSE are reported.
}
\label{ablation_study_vsensenet_1}
\end{table*}
}

\subsection*{S.1.2 VSenseNet II}

\begin{figure*}
    \centering
    \includegraphics[width=17cm]{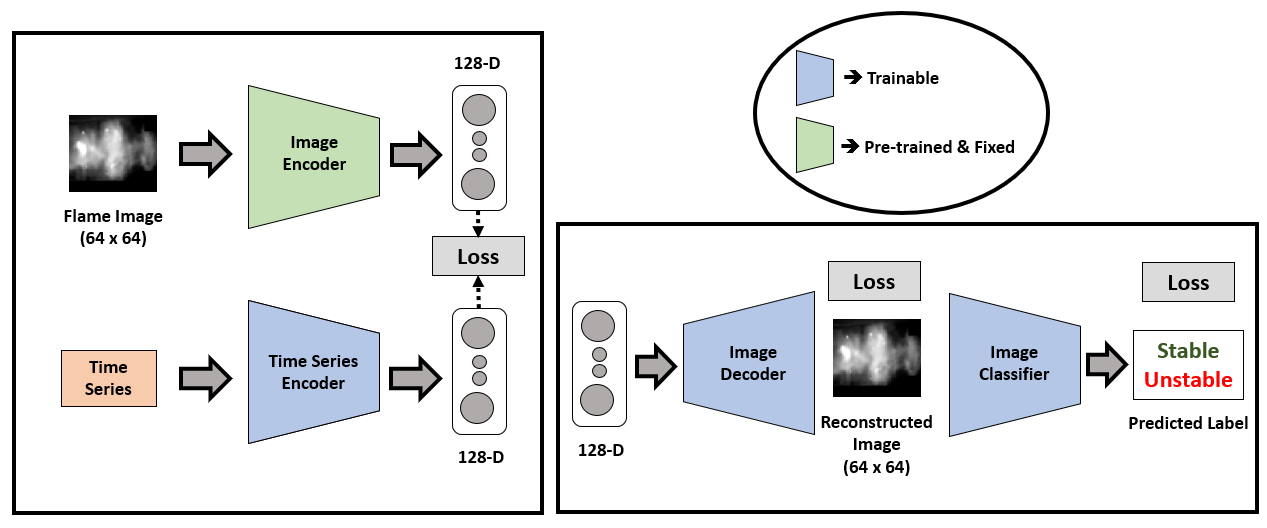}
    \caption{Training framework of VSenseNet II. It comprises two steps. The first step is to train the time series encoder. The next step is to train the image decoder and image classifier utilizing reconstruction loss and classification loss.}
    \label{vsensenet_2_training}
\end{figure*}

\begin{figure*}
    \centering
    \includegraphics[width=17cm]{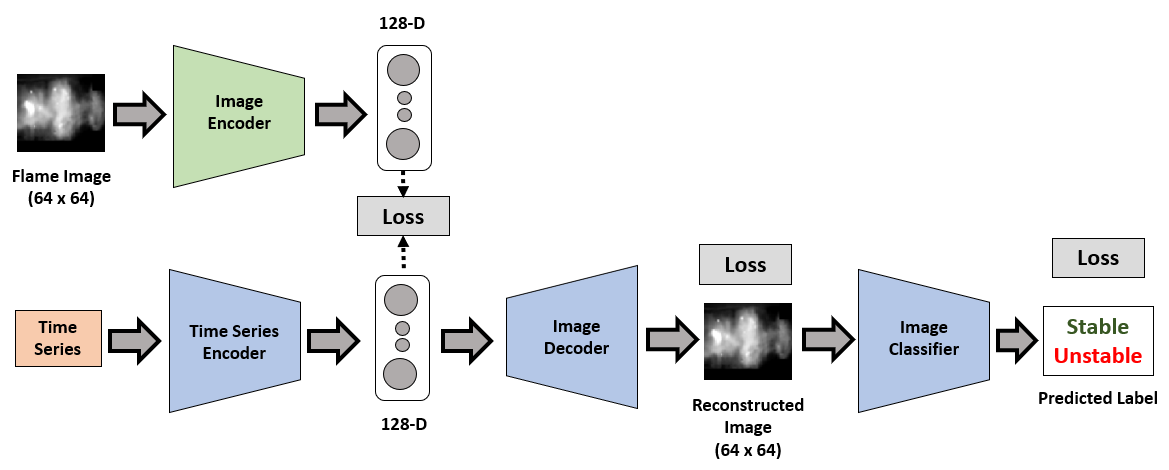}
    \caption{Training framework of VSenseNet II(A). It consists of a single step to train the time series encoder, image decoder and image classifier.}
    \label{vsensenet_2A_training}
\end{figure*}

Compared to VSenseNet I, VSenseNet II has the image classifier model in the training framework apart from the time series encoder and the image decoder. We demonstrate the training framework of VSenseNet II in Fig.~\ref{vsensenet_2_training}. The trainable parts are the time series encoder, image decoder, and image classifier. 
With the time series as input, the time series encoder computes a 128-dimensional time series embedding.
From a time series embedding, the image decoder learns to reconstruct the corresponding flame image as closely as possible. With the reconstructed flame image as input, the image classifier model predicts it as stable or unstable.

As part of the ablation study for VSenseNet II, we develop the model VSenseNet II(A) demonstrated in Fig.~\ref{vsensenet_2A_training}.
While VSenseNet II consists of two steps for training, VSenseNet II(A) consists of a single step to train the time series encoder, image decoder, and image classifier. From Table~\ref{ablation_study_vsensenet_2}, we observe that in terms of classification performance VSenseNet II is better than VSenseNet II(A). For reconstruction, VSenseNet II(A) performs better, but the performance is not as good as VSenseNet I. 

{
\setlength{\tabcolsep}{6pt}
\begin{table*}[h]
\begin{center}
    \begin{tabular}{|c||c|c||c|c|c|}
    \hline
    \multirow{2}{*}{Model} &
    \multicolumn{2}{c||}{Reconstruction Performance} & \multicolumn{3}{c|}{Classification Performance} \\ 
    \cline{2-3} \cline{4-5} \cline{5-6}
    & SSIM & Mean Squared Error & Accuracy & F1 Score & FNR\\
    \hline
    \hline
    Image Classifier & NA & NA & \textbf{0.9938 $\pm$ 0.0064} & \textbf{0.9938 $\pm$ 0.0065} & \textbf{0.0122 $\pm$ 0.0129}\\
    \hline
    Time Series Classifier & NA & NA & \textbf{0.9850 $\pm$ 0.0021} & \textbf{0.9847 $\pm$ 0.0022} & \textbf{0.0300 $\pm$ 0.0044}\\
    \hline
    \hline
    Cross-Modal & 0.6655 & 0.0072 & 0.9888 $\pm$ 0.0005 & 0.9887 $\pm$ 0.0005 & 0.0222 $\pm$ 0.0010\\
    \hline
    VSenseNet II & 0.6876 & 0.0070 & \textbf{0.9901 $\pm$ 0.0003} & \textbf{0.9900 $\pm$ 0.0003} & \textbf{0.0197 $\pm$ 0.0007}\\
    \hline
    VSenseNet II(A) & \textbf{0.6978} & \textbf{0.0065} & 0.9898 $\pm$ 0.0006 & 0.9897 $\pm$ 0.0006 & 0.0202 $\pm$ 0.0012\\
    \hline
    \end{tabular}
\end{center}
\caption{Ablation study with VSenseNet II for the test set. For classification performance, average and standard deviation of the evaluation metrics (Accuracy, F1 Score, FNR) are reported after training each model five times. For reconstruction performance, average values of SSIM and MSE are reported.
}
\label{ablation_study_vsensenet_2}
\end{table*}
}

\section*{S.2 Convolutional Autoencoder}

\begin{figure*}
    \centering
    \includegraphics[width=12cm]{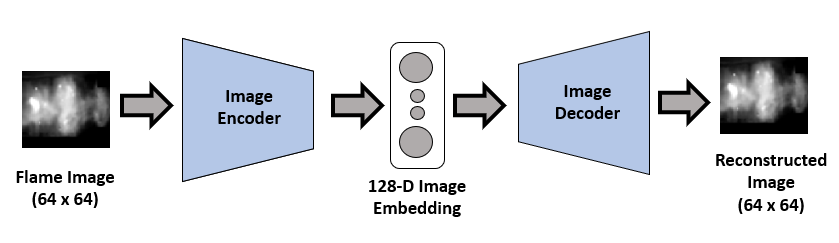}
    \caption{The encoder and decoder of the convolutional autoencoder model used for pre-training.}
    \label{autoencoder}
\end{figure*}

\begin{figure*}
    \centering
    \includegraphics[width=16cm]{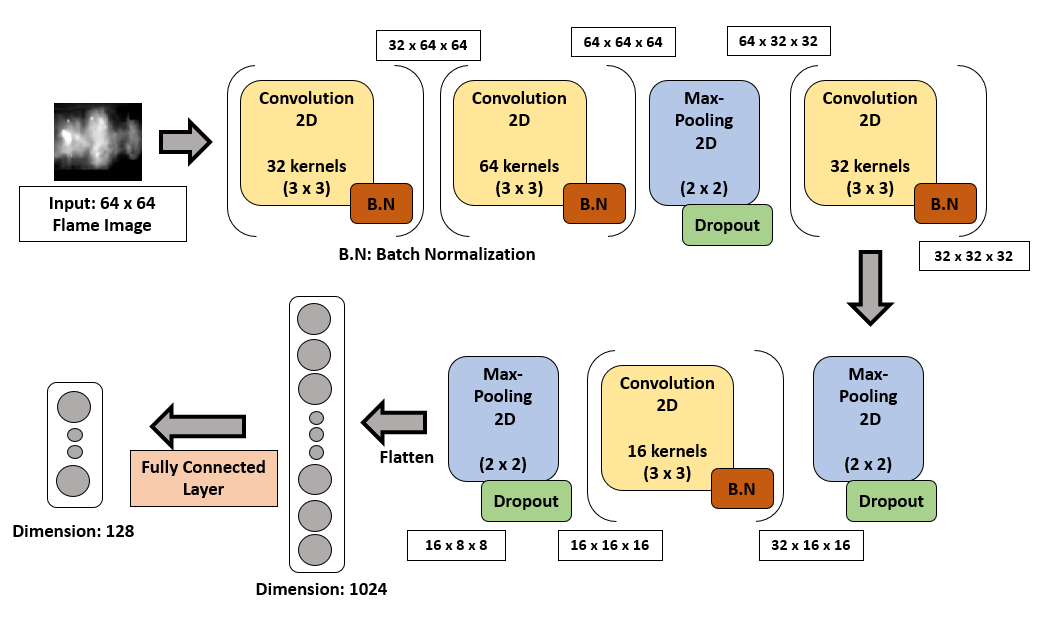}
    \caption{Encoder model for pre-training of convolutional autoencoder.}
    \label{encoder}
\end{figure*}

\begin{figure*}
    \centering
    \includegraphics[width=16cm]{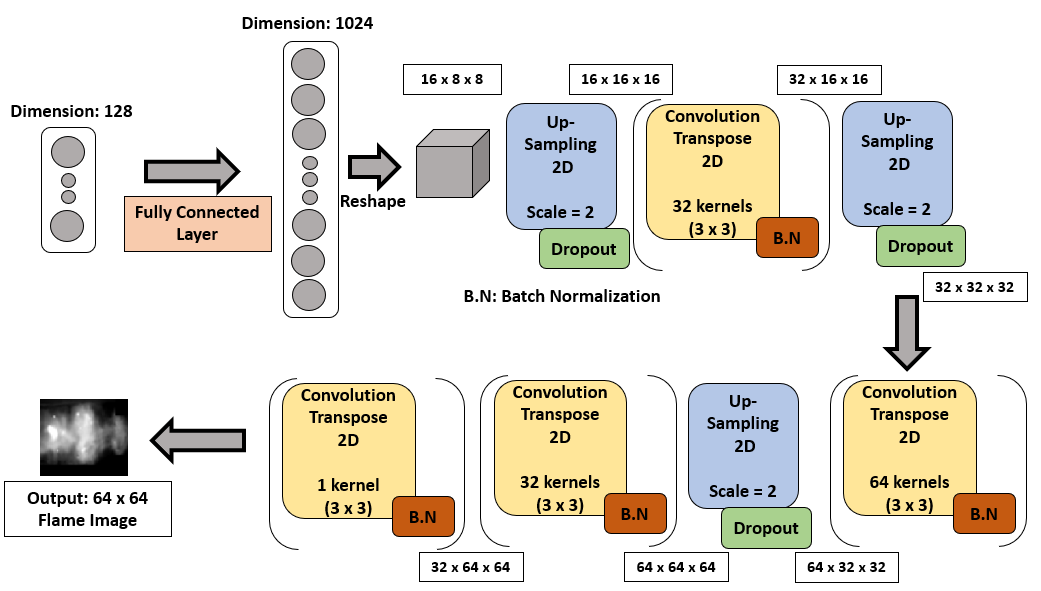}
    \caption{Decoder model for pre-training of convolutional autoencoder.}
    \label{decoder}
\end{figure*}

Autoencoders can learn meaningful representations using a compression function (encoder) and a decompression function (decoder). The encoder compresses the input into a low dimensional embedding, and the decoder reconstructs the high dimensional information from that embedding.
Without the requirement of explicit annotations, the weights of an autoencoder model can be learned with the objective of minimizing the reconstruction loss.
The first step of developing our proposed framework is to utilize the training dataset of flame images to pre-train a convolutional autoencoder which comprises an image encoder and an image decoder as demonstrated in Fig.~\ref{autoencoder}.
The encoder takes in a flame image (resolution 64 x 64) as input. The encoder model comprises a series of 2D convolutional and 2D max-pooling layers. After that, a fully connected layer is used to compute a 128-dimensional embedding. From the 128-dimensional embedding, the decoder attempts to reconstruct the original flame image as closely as possible using a series of 2D up-sampling and 2D convolutional transpose layers. The details of the image encoder and decoder model are shown in Fig.~\ref{encoder} and Fig.~\ref{decoder} respectively.

\section*{S.3 Baseline Models}

\subsection*{S.3.1 Image Classifier}

\begin{figure*}
    \centering
    \includegraphics[width=16cm]{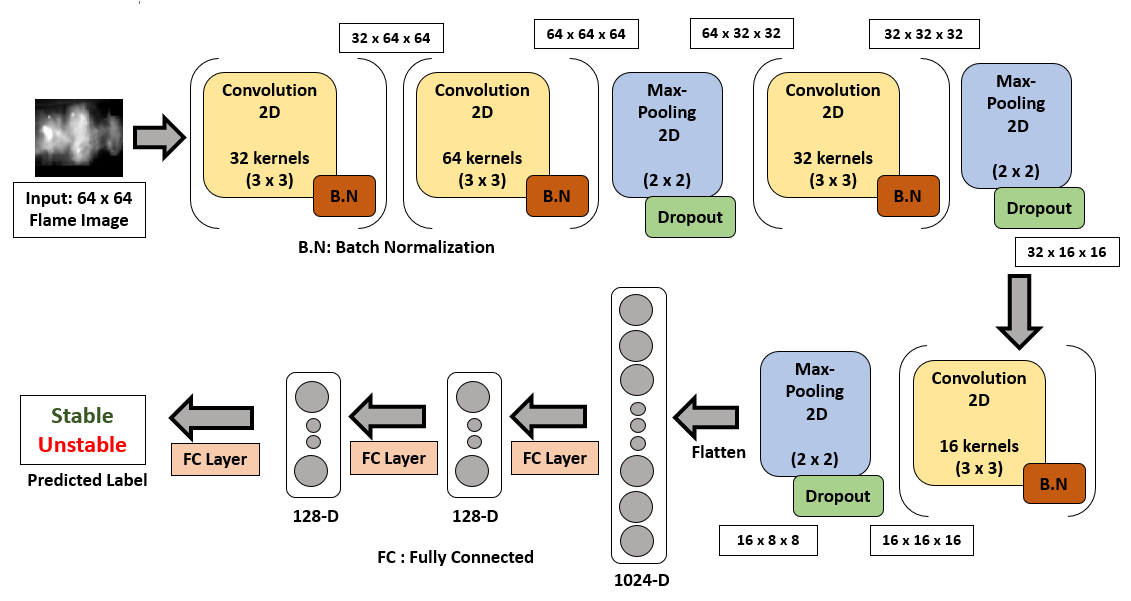}
    \caption{Image Classifier model classifies a flame image as stable or unstable.}
    \label{image_classifier}
\end{figure*}

This image-based model is a binary classification model which takes a flame image as input and classifies it as stable or unstable. It is trained on actual training set images and tested on actual test set images. The image classifier model comprises 2D convolutional, 2D max-pooling, and fully connected layers. The model architecture of the image classifier is shown in Fig.~\ref{image_classifier}.

\subsection*{S.3.2 Time Series Classifier}

\begin{figure*}
    \centering
    \includegraphics[width=16cm]{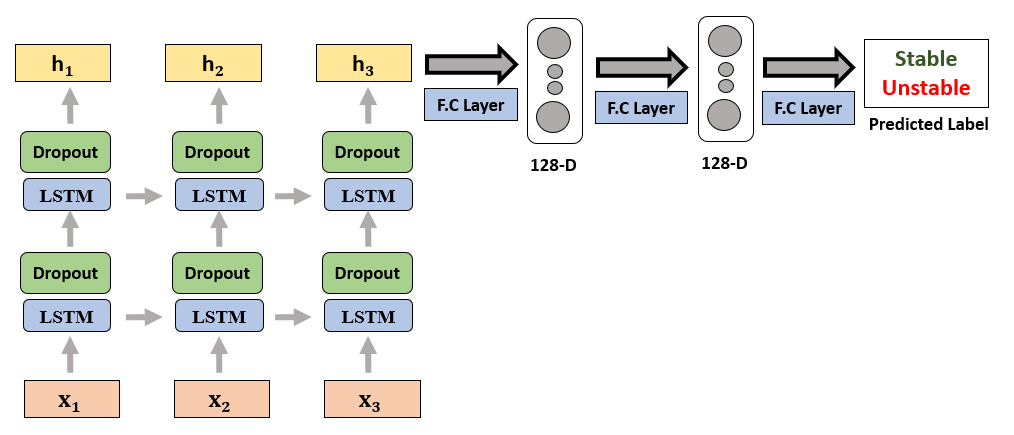}
    \caption{Time Series Classifier model classifies an acoustic pressure time series as stable or unstable.}
    \label{ts_classifier}
\end{figure*}

This is an entirely time series based model with no cross-modal reconstruction of visual features. The model architecture of the time series classifier has been demonstrated in Fig.~\ref{ts_classifier}. With the multivariate time series as input, the model predicts it as stable or unstable. The time series classifier model is trained using time series only, and it is tested using time series of the test set.

\section*{S.4 Dataset Collection}

\begin{figure*}
    \centering
    \includegraphics[width=14cm]{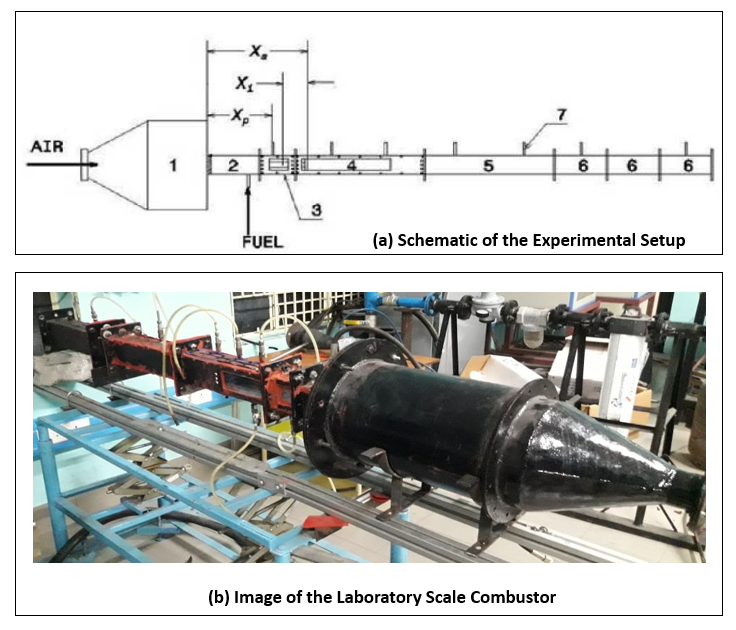}
    \caption{The laboratory-scale combustor used for data collection. (a) Schematic of the experimental setup. 1 - settling chamber, 2 - inlet duct, 3 - IOAM, 4 - test section, 5 - big extension duct, 6 – small extension ducts, 7 - pressure transducers, Xs - swirler location measured downstream from settling chamber exit, Xp - transducer port location measured downstream from settling chamber exit, Xi - fuel injection location measured upstream from swirler exit. (b) Image of the laboratory scale combustor.}
    \label{experimental_setup}
\end{figure*}

For dataset collection, we induce combustion instability in a laboratory-scale swirl combustor (Fig.~\ref{experimental_setup}), which has a swirler of diameter 30 mm and vane angles of 60 degrees. Air is provided to the combustor through a settling chamber of diameter 28 cm and thereafter through a square cross-section of side 6 cm. The experimental setup includes an inlet section, an inlet optical access module (IOAM), a primary combustion chamber, and a secondary duct. The IOAM facilitates optical access to the fuel tube. The fuel is injected co-axially with air through a fuel injection tube, having slots on the surface at selected distances upstream of the swirler.

The chosen upstream distances are 90 mm and 120 mm. For the upstream distance of 90 mm, partial premixing of the fuel with air occurs, while the distance of 120 mm facilitates full premixing of the fuel and air. 
Based on the swirler diameter, different airflow rates are chosen for a fixed fuel flow rate. In another way, the inlet air flow rate can be kept fixed for different fuel flow rates.

\end{document}